\documentclass{article}

\usepackage{arxiv}
\usepackage{amsmath}
\usepackage{subcaption}
\usepackage[utf8]{inputenc} 
\usepackage[T1]{fontenc}    
\usepackage{hyperref}       
\usepackage{url}            
\usepackage{booktabs}       
\usepackage{amsfonts}       
\usepackage{nicefrac}       
\usepackage{microtype}      
\usepackage{lipsum}		
\usepackage{graphicx}
\usepackage{natbib}
\usepackage{doi}

\title{Restoring Neural Network Plasticity for Faster Transfer Learning}

\date{} 					

\author{ \href{https://orcid.org/0009-0003-5495-5355}{\includegraphics[scale=0.06]{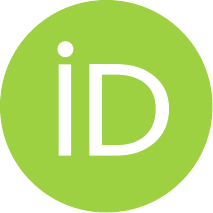}\hspace{1mm}Xander Coetzer} \\
	Department of Computer Science\\
	University of Pretoria\\
	Pretoria, South Africa \\
	\texttt{u20465026@tuks.co.za} \\
	\And
	\href{https://orcid.org/0000-0002-9061-1141}{\includegraphics[scale=0.06]{orcid.pdf}\hspace{1mm}Arn\'{e} Schreuder} \\
	Department of Computer Science\\
	University of Pretoria\\
	Pretoria, South Africa \\
	\texttt{arneschreuder@me.com} \\
	\AND
	\href{https://orcid.org/0000-0003-3546-1467}{\includegraphics[scale=0.06]{orcid.pdf}\hspace{1mm}Anna Sergeevna Bosman} \\
	Department of Computer Science\\
	University of Pretoria\\
	Pretoria, South Africa \\
	\texttt{anna.bosman@up.ac.za} \\
}

\renewcommand{\shorttitle}{Restoring Neural Network Plasticity for Faster Transfer Learning}

\hypersetup{
pdftitle={Restoring Neural Network Plasticity for Faster Transfer Learning},
pdfauthor={Xander Coetzer, Arn\'{e} Schreuder, Anna S. Bosman},
pdfkeywords={plasticity, neural networks, transfer learning},
}

\begin{document}
\maketitle

\begin{abstract}
	Transfer learning with models pretrained on ImageNet has become a standard practice in computer vision. Transfer learning refers to fine-tuning pretrained weights of a neural network on a downstream task, typically unrelated to ImageNet. However, pretrained weights can become saturated and may yield insignificant gradients, failing to adapt to the downstream task. This hinders the ability of the model to train effectively, and is commonly referred to as loss of neural plasticity. Loss of plasticity may prevent the model from fully adapting to the target domain, especially when the downstream dataset is atypical in nature. While this issue has been widely explored in continual learning, it remains relatively understudied in the context of transfer learning. In this work, we propose the use of a targeted weight re-initialization strategy to restore neural plasticity prior to fine-tuning. Our experiments show that both convolutional neural networks (CNNs) and vision transformers (ViTs) benefit from this approach, yielding higher test accuracy with faster convergence on several image classification benchmarks. Our method introduces negligible computational overhead and is compatible with common transfer learning pipelines.
\end{abstract}


\section{Introduction}
Transfer learning has become a cornerstone of modern computer vision workflows, largely due to the widespread availability of powerful deep learning models pretrained on large-scale datasets such as ImageNet~\citeyearpar{geetha2017review}. In typical scenarios, these pretrained models are fine-tuned on a downstream dataset that often differs substantially in size, structure, or domain compared to the dataset used for pretraining. While the transfer learning approach offers significant gains in both performance and data efficiency, it also introduces critical challenges that remain underexplored.

One such challenge is the loss of neural plasticity, i.e., the reduced ability of a neural network to adapt to new data distributions during fine-tuning. This issue arises when certain pretrained weights, particularly those with near-zero magnitudes, result in negligible gradients during backpropagation~\citeyearpar{lyle2022understanding}. Such weights become effectively inactive, hindering learning and preventing the model from fully adapting to the target domain. Although loss of neural plasticity has been extensively studied in the context of continual learning~\citeyearpar{dohare2024loss,lyle2023understanding}, it has received limited attention in transfer learning, where similar dynamics can emerge due to the abrupt domain shift between pretraining and fine-tuning datasets~\citeyearpar{magotra2019transfer}.

Recent research in continual learning proposed a selective weight reinitialization strategy to restore plasticity by resetting weights deemed unimportant or inactive~\citeyearpar{hernandezgarcia2025reinitializingweightsvsunits}. Inspired by this line of work, our research investigates the application of weight reinitialization techniques within the transfer learning context. We selectively reinitialize a subset of low utility pretrained weights prior to fine-tuning to restore their capacity for learning. We evaluate the effectiveness of this approach across multiple architectures, including convolutional neural networks (CNNs) and vision transformers (ViTs), on a range of datasets of varying size and complexity. Novel contributions of this work are summarized as follows:
\begin{itemize}
    \item We establish the relationship between neural plasticity and transfer learning.
    \item Neural plasticity is restored prior to fine-tuning through a targeted weight reinitialization strategy, which identifies and reinitializes inactive weights.
    \item The proposed approach improves downstream classification accuracy, especially in tasks involving substantial domain shifts.
    \item Further, our method enhances training efficiency by enabling faster convergence, without increasing model size or requiring architectural modifications.
\end{itemize}
By addressing a critical but underexplored aspect of transfer learning, our work provides a principled way to improve model adaptation and performance on diverse downstream tasks. This method is lightweight, easily integrated into existing pipelines, and generalizes well across model types and datasets.

The rest of the paper is structured as follows: Section~\ref{sec:background} discusses the essential background topics and related work, Section~\ref{sec:methodology} details the proposed methodology and experimental setup. Section~\ref{sec:results} presents the experimental results, and Section~\ref{sec:analysis} critically analyses the findings. Finally, Section~\ref{sec:conclusion} concludes the paper and offers some directions for future research.

\section{Background and Related Work}\label{sec:background}
To contextualize the proposed approach, the following related topics are discussed in this section: transfer learning (Section~\ref{sec:transfer}), continual learning and neural plasticity (Section~\ref{sec:plasticity}), weight reinitialization (Section~\ref{sec:background:reinitialization}), and parameter efficient fine-tuning (Section~\ref{sec:related}).
\subsection{Transfer Learning}\label{sec:transfer}
Transfer learning is a technique that aims to transfer knowledge from one model to another~\citeyearpar{sital20203d}. Practically, this involves using the weights that have been learned by the former model as a starting point for training the new model~\citeyearpar{geetha2017review}. It is especially useful in areas where the amount of available, labeled training data is limited. The performance of the new model scales with the size of the source dataset as well as the similarity between the source and target datasets~\citeyearpar{reed2022self}. The transition from the source distribution to the target distribution can cause loss of neural plasticity \citeyearpar{magotra2019transfer}, similar to what is observed in continual learning.

\subsection{Continual Learning and Neural Plasticity} \label{sec:plasticity}
Continual learning refers to the scenario in which a model is trained on a dynamic data distribution, i.e., the data distribution changes as training progresses. Continual learning in neural networks is enabled via neural plasticity and memory stability. The former refers to the ability of the model to adapt well to new data, while the latter refers to the ability of the model to remain performant on old data~\citeyearpar{wang2024comprehensive}. Transfer learning has no need for memory stability, but is similar to continual learning as far as neural plasticity is concerned. 

Plasticity is a term adopted from neuroscience, and refers to the ability of a neural network model to adapt its weights when trained on new data which the model has not encountered before~\citeyearpar{lyle2023understanding}. Loss of plasticity occurs when a model being trained on new data cannot ``overwrite'' previously learned patterns to properly fit the new distribution~\citeyearpar{lyle2022understanding}. Dohare et al.~\citeyearpar{dohare2024loss} introduce three main causes of loss of plasticity in continual learning. The first is an increase in constant or dead units, which occurs when gradients become insignificant and weights are no longer updated to effectively fit the downstream task. The second is a steady growth of the average weight magnitude, as larger weights often lead to slower training. Lastly, loss of plasticity can be caused by a drop in the effective rank of the representation. Such drop occurs when the output is determined by a subset of the weights in the network, rather than each weight having a more or less equal influence. This paper will focus on addressing the first of these causes, i.e., non-contributing weights. Since small weights are less likely to significantly impact the output of the network~\citeyearpar{hernandezgarcia2025reinitializingweightsvsunits}, we focus our study on pretrained weights close to zero.

\subsection{Reinitialization Algorithms}\label{sec:background:reinitialization}
Reinitialization algorithms have been effectively applied to increase generalization performance in neural networks and maintain plasticity in the context of continual learning \citeyearpar{dohare2024loss,nikishin2022primacy,sokar2023dormant}. Most notably, Hernandez-Garcia et al.~\citeyearpar{hernandezgarcia2025reinitializingweightsvsunits} proposed a selective weight reinitialization approach, focusing on modifying specific weights within a neural network rather than entire units or layers. Their method comprises three components: a utility function for evaluating weight importance, a pruning function to determine which weights to reset, and a reinitialization strategy to assign new values to the pruned weights.
\paragraph{Utility Functions:}
The utility function acts as a proxy to identify the least useful weights. Hernandez-Garcia et al.~\citeyearpar{hernandezgarcia2025reinitializingweightsvsunits} consider two utility metrics: \textit{magnitude-based} and \textit{gradient-based}. Formally, let $W$ denote the set of all weights in the model, and let $g_w$ denote the derivative loss with respect to a weight $w$. Then for each weight $w\in W$ the magnitude utility returns $|w|$ and the gradient utility returns $|w\cdot g_w|$. 
\paragraph{Pruning Functions:}
The pruning function determines which weights to reset. Hernandez-Garcia et al.~\citeyearpar{hernandezgarcia2025reinitializingweightsvsunits} define two pruning functions. The first is \textit{proportional pruning}, which selects a proportion $k \in (0,1]$ of the weights. If the number of weights to reinitialize is not an integer, it is converted as follows: Take the integer part of the real number and add to it $b$ where $b$ is the result of sampling from a Bernoulli distribution with probability of success equal to the decimal part of the real number. The second is \textit{threshold pruning}, which defines a fixed threshold value, and reinitializes the weights with absolute values below the threshold.
\paragraph{Reinitialization Methods:}
The reinitialization method determines the new value of the pruned weight. Hernandez-Garcia et al.~\citeyearpar{hernandezgarcia2025reinitializingweightsvsunits} discuss two reinitialization methods, namely, \textit{resample reinitialization} and \textit{mean reinitialization}. \textit{Resample reinitialization} reinitializes the weight by sampling from the initialization distribution, and \textit{mean reinitialization} reinitializes to the mean of the initial distribution.


\subsection{Related Work: Parameter Efficient Tuning}\label{sec:related}
As a result of the widespread use of increasingly large foundation models \citeyearpar{bommasani2021opportunities}, the need to improve the efficiency of fine-tuning becomes increasingly evident. The field of parameter-efficient fine-tuning (PEFT) is a study of how this can be achieved \citeyearpar{han2024parameter}. The following section highlights works of interest as they relate to the results presented in this paper.

\paragraph{Guided Training:}The fine-tuning process can be guided with the goal of improving transfer learning accuracy. A regularization term can be defined to penalize the model for drifting away from the pretrained weights during fine-tuning~\citeyearpar{xuhong2018explicit}. A similar approach is to only apply strong regularization to layers that show inconsistent loss reduction, and weaker regularization to layers where the pretrained weights are more useful~\citeyearpar{tian2024rethinking}. Although our method similarly aims to enhance fine-tuning accuracy through targeted adjustments to specific weight layers, it distinguishes itself by prioritizing the restoration of learning plasticity through reinitialization of weights before training, rather than guided weight updates during training.

\paragraph{Weight Reduction:}
Fine-tuning efficiency can be improved by reducing the number of weights which have to be adjusted. All pretrained weights can be frozen, allowing only newly added layers to be trained during fine-tuning~\citeyearpar{chen2024conv}. Other approaches include Batchnorm tuning~\citeyearpar{mudrakarta2018k} and Bias tuning~\citeyearpar{zaken2021bitfit}, where all layers in the model are frozen during fine-tuning except the Batchnorm and Bias layers, respectively. All of these only update partial parameters of the network, unlike our method where no weights are frozen and all layers can be updated during the fine-tuning step. 

\paragraph{Summary:}
Existing approaches to improve transfer learning focus mostly on the learning task (how should the loss be formulated, what penalty terms can be defined to guide learning) and parameter efficiency (what parts of the architecture can be frozen and not take part in fine-tuning). In both cases, the ability of the neural network to learn and adapt to the new task is presumed. We argue that such assumption is not necessarily valid, and investigate the effectiveness of restoring neural plasticity as a method of improving fine-tuning efficacy. A secondary objective of our research is to increase the usefulness of transfer learning when fine-tuning on atypical datasets that may be poorly aligned with the pretrained dataset.

\section{Methodology}\label{sec:methodology}

This study evaluates the application of weight reinitialization (Section \ref{sec:approach}) to reduce the impact of plasticity loss on CNNs and ViTs in the context of transfer learning. An empirical approach is taken in this paper, where experiments are performed on a variety of image classification benchmarks (Section \ref{sec:datasets}) and architectures (Section \ref{sec:architecture}). 

\subsection{Approach}\label{sec:approach}
The proposed method is based on the selective weight reinitialization algorithm discussed in Section \ref{sec:background:reinitialization}. The continual version of this algorithm is applied every certain number of epochs \citeyearpar{hernandezgarcia2025reinitializingweightsvsunits}. In the transfer learning context, selective weight reinitialization is applied only once, before fine-tuning. In all cases, biases are reset to zero at the start of fine-tuning to give the model a neutral starting point on the new task. The rest of this section details the key components of the approach, namely, the utility, pruning, and reinitialization functions. 

\paragraph{Utility Functions:}
In a recent study, van der Grijp et al. \citeyearpar{henri2025visualizing} showed that initial gradient is not a reliable indicator for weight saliency. Since we only apply reinitialization at the beginning of fine-tuning, we focus our attention on the magnitude utility, which is used as defined in Section \ref{sec:background:reinitialization}.

\paragraph{Pruning Functions:}
We use the proportional pruning algorithm, but instead of sampling from a Bernoulli distribution when a real value is encountered, the proportion of non-contributing weights is rounded down. In order to study the sensitivity of models to different degrees of change, we experiment with three fixed pruning ratios, namely 5\%, 10\%, and 25\% of total weights.

\paragraph{Reinitialization Functions:}
Four different reinitialization methods were considered. Firstly, \textit{mean reinitialization} was adapted to reset stagnant weights to the mean of the pretrained distribution instead of the initial distribution in order to maintain the useful features already learned. Second variation added controlled noise to mean reinitialization, sampled from a normal distribution with mean $\mu= 0$ and standard deviation $\sigma = 1$ prior to being scaled to an order of magnitude less than the remaining weights within the layer. Noise generation is formalized as:
\begin{equation}
noise=\frac{N(0,1)\cdot \mu}{10}
\end{equation}
This method seeks to balance stability with the benefits of stochasticity, inspired by prior work suggesting that mild noise can enhance generalization \citeyearpar{dohare2024loss}.

The final two variants are both adaptions of resample reinitialization. The third variant samples directly from a Normal distribution with $\mu= 0$ and $\sigma = 0.2$, and  the fourth variant samples from a Normal distribution with  $\mu= 0$ and $\sigma = 1$. The third variant is designed to mimic the weight distribution observed in the models after pretraining, and the fourth was included to evaluate the effect of high magnitude changes on the final fine-tuning results.

To concisely reference specific experimental configurations, we introduce the notation 
\begin{equation}
\langle P\rangle \langle RF\rangle \text{ where } P \in \{5,10,25\} \text{ and } RF \in \{ M,MN,NS,N\}
\end{equation}
Here, $P$ is the percentage parameter that specifies the proportion of weights to be reset, as discussed above, and $RF$ is the reinitialization function. $M$ denotes mean reinitialization, $MN$ denotes mean reinitialization with noise, $NS$ denotes reinitialization by sampling from Normal$(0,0.2)$, and $N$ denotes reinitialization by sampling from Normal$(0,1)$.

 \subsection{Datasets}\label{sec:datasets}
 To assess the effectiveness of the re-initialization algorithms, we conduct experiments on three small datasets varying in domain. Table~\ref{tab:Datasets} summarizes the overall class and sample statistics per dataset. Each individual dataset is discussed below.

\begin{table}[!htb] 
\centering
\caption{The number of classes, number of images per class and total number of images within each dataset. }\label{tab:Datasets}
\begin{tabular}{|l|l|l|l|}
\hline
Dataset & Number of Classes &  Images per Class & Total Images\\
\hline
DTD & 47 & 120 & 5640 \\
Brain Tumor & 4 & 1600-2000 & 7150 \\
Fruit25 & 25 & 450 & 11250 \\
\hline
\end{tabular}
\end{table}
 
 \paragraph{DTD:}The Describable Textures Dataset (DTD) \citeyearpar{cimpoi14describing} emphasizes low-level, tex-ture-based descriptors (e.g., ``bubbly,'' ``striped'') rather than object semantics. In this context, object semantics refers to the high-level meaning or category identity of an object (e.g., “dog,” “apple”) which are features that are central to ImageNet classification but absent in DTD. 
 
 \paragraph{Brain Tumor:}The Brain Tumor dataset \citeyearpar{brain2024} is an atypical medical magnetic resonance imaging (MRI) dataset. Unlike ImageNet’s natural photographs, MRI scans differ substantially in modality, texture statistics, and visual distribution, making them semantically and visually distinct. 
 
 \paragraph{Fruit25:} The Fruit25 dataset is a subset of Fruit100 \citeyearpar{Fruit100}, and includes 25 randomly selected categories of fruit. This subset selection was made to ensure manageable training time while preserving sufficient class diversity.  The dataset contains natural images (photographs) of fruit.\\
 
 The object semantics and visual characteristics of Fruit25 are more closely aligned with ImageNet than that of DTD and Brain Tumor. This enables us to directly compare performance between datasets that are semantically similar to the source distribution and those that are not, thereby isolating the effect of domain similarity on plasticity and generalization.

\subsection{Architecture}\label{sec:architecture}

To assess the influence of model type and capacity on the effectiveness of selective weight reinitialization, we conduct experiments using architectures of varying complexity, listed in Table \ref{tab:Architecture}. Specifically, we employ SimpleNet \citeyearpar{hasanpour2016lets} and ResNet-50 \citeyearpar{he2016deep}, both of which are CNNs of varying sizes. To evaluate the generalizability of our approach beyond CNNs, we also include the ViT-B16 model, which was introduced as one of the first ViT models by Dosovitskiy et al. \citeyearpar{dosovitskiy2020image}. This diverse architectural selection enables us to examine the consistency of reinitialization benefits across distinct architectures and parameter scales.

\begin{table}[!htb] 
\centering
\caption{Total number of trainable parameters per model.}\label{tab:Architecture}
\begin{tabular}{|l|l|}
\hline
Model & Number of Parameters\\
\hline
SimpleNet & ~ 0.5 million \\
ResNet-50 & ~ 25.6 million \\
ViT-B16   & ~ 86.6 million \\
\hline
\end{tabular}
\end{table}

\subsection{Experimental Setup}\label{sec:experimental_setup}
Each experiment was repeated ten times over multiple independent and identical runs to account for variability due to random initialization and training dynamics. To mitigate overfitting, we employ early stopping \citeyearpar{dishar2023review} with a patience threshold of 10 epochs. For each experimental case, we report the mean and standard deviation (std) of the test accuracy, as well as the mean and std for the number of epochs it took the model to converge.

The base case consists of a model pretrained on ImageNet, reflecting the prevailing standard in transfer learning for large-scale image classification tasks~\citeyearpar{chen2024conv,xuhong2018explicit}. The proposed method modifies this baseline by selectively reinitializing a subset of the model's weights prior to fine-tuning. We evaluate three model architectures across three different datasets and four reinitialisation settings. No hyperparameter tuning is conducted; all models use default parameters as provided by their existing PyTorch implementation. The source code and instructions for reproducing all experiments are publicly available on \href{https://github.com/Xander-FP/Restoring-Neural-Network-Plasticity-for-Faster-Transfer-Learning}{GitHub}.

\section{Empirical Results}\label{sec:results}
This section presents empirical results, evaluating different experimental cases across three distinct architectures. The results are grouped by architecture: SimpleNet (Section~\ref{sec:simple}), ResNet-50 (Section~\ref{sec:resnet}), and ViT-B16 (Section~\ref{sec:vit}). The notation for each experimental case is used as defined in Section \ref{sec:approach}. Statistical significance testing is performed to compare the base case with the top experimental run for each dataset and architecture pair. Since the distribution of the experimental data is not known, the Mann Whitney U test~\citeyearpar{mcknight2010mann} is used to ensure the statistical significance of the results. The Mann-Whitney U test is a nonparametric test of the null hypothesis that does not make assumptions regarding the distribution of the data. For our purposes, the null hypothesis is that the best experimental case for each dataset and architecture pair and the base case are from the same distribution, while the alternative hypothesis is that the best experimental case for each dataset and architecture pair comes from a distribution with a larger mean than that of the base case distribution. 

\subsection{SimpleNet}\label{sec:simple}
 
Results for SimpleNet, the smallest architecture considered, are summarized in Table~\ref{tab:SimpleNet}. Table~\ref{tab:SimpleNet} compares the base case with the top three results for each dataset. Within each dataset, the various experimental cases are sorted in descending order in terms of average accuracy. As such, the best performing case appears at the top of each group. The baseline is indicated in bold. 

On the DTD dataset, all $M$ and $MN$ experimental cases outperformed the base case, with the largest accuracy gain reaching approximately 2\% for the 10 $M$ experimental case, with no notable change in convergence rate. The p-value for comparing the accuracy gained by the top performing 10$M$ experimental case is $0.03836$, indicating a significant improvement on the base case for $p<0.05$. For the Brain Tumor dataset, the base case achieved the highest accuracy, maintaining roughly 1\% lead over the top plasticity-restoring variants. However, the best-performing plasticity configurations converged in an average of $100$ epochs, a substantial efficiency improvement when compared to $170$ epochs for the base case. The distributions of the base case and the next best experimental case, 10 $MN$ is now compared. The alternative hypothesis for the accuracy comparison is adjusted for this comparison such that it tests whether the distribution of the base case has a larger mean than the distribution of the 10 $MN$ case. The resultant p-value for the accuracy distribution comparison is $0.2109$. The p-value comparing the distributions of the epochs is $0.09342$. In both cases the difference is not significant at $p<0.05$. On the Fruit dataset, two experimental cases demonstrated improvements in both accuracy and rate of convergence, with a maximum accuracy increase of 0.5\% and a convergence speedup of 18 epochs. Neither of these are significant at $p<0.05$ with p-values of $0.2946$ and $0.46812$, respectively, for the comparison of the distribution of the accuracy and the convergence speedup between the base case and the 10 $M$ experimental case.

Overall, only the accuracy improvement for the 10$M$ case is significant at $p<0.05$. There is also no instance where the base case significantly outperforms the strongest SimpleNet plasticity-restoring configuration.

\begin{table}[!htb]
\centering
\caption{A comparison between plasticity-restoring experimental cases and the base case on the \textit{SimpleNet} model. The specific experimental cases are listed using the notation defined in Section \ref{sec:approach}. The results are ordered by accuracy, and the base case is indicated in bold.}\label{tab:SimpleNet}
\begin{tabular}{|c|c|l@{\ $\pm$\,}l|l@{\ $\pm$\,}r|}
\hline
Dataset & Experimental Case &  Accuracy & Std & Epochs & Std\\
\hline
& 10 M & 60.355 & 1.148 & 96.000 & 9.110 \\
DTD & 10 MN & 59.734 & 1.684 & 88.000 & 20.736 \\
 & 5 MN & 59.220 & 0.251 & 88.500 & 6.364 \\
& \textbf{Base} & \textbf{57.926} & \textbf{1.463} & \textbf{89.400} & \textbf{5.899 }\\
 \hline
& \textbf{Base} & \textbf{96.087} & \textbf{0.781} & \textbf{169.167} & \textbf{61.411} \\
Brain Tumor & 10 MN & 95.304 & 0.583 & 107.200 & 18.102 \\
 & 10 M & 94.885 & 1.355 & 85.600 & 33.471 \\
& 5 M & 94.822 & 1.136 & 89.364 & 35.192 \\
  \hline
& 10 M & 82.720 & 0.226 & 100.667 & 41.284 \\
Fruit 25 & 25 M & 82.320 & 0.905 & 100.333 & 57.143 \\
& \textbf{Base} & \textbf{82.256} & \textbf{0.694} & \textbf{118.600} & \textbf{12.178} \\
& 5 MN & 82.120 & 0.057 & 84.333 & 59.282 \\
\hline
\end{tabular}
\end{table}

\subsection{ResNet-50}\label{sec:resnet}
Results for ResNet-50, a deeper and more complex CNN, are summarized in Table~\ref{tab:ResNet-50}.

\begin{table}[!htb]
\centering
\caption{A comparison between plasticity-restoring experimental cases and the base case on the \textit{ResNet-50} model. The specific experimental cases are listed using the notation defined in Section \ref{sec:approach}. The results are ordered by accuracy, and the base case is indicated in bold.}\label{tab:ResNet-50}
\begin{tabular}{|c|c|l@{\ $\pm$\,}l|l@{\ $\pm$\,}r|}
\hline
Dataset & Experimental Case &  Accuracy & Std & Epochs & Std\\
 \hline
& 5 M & 72.030 & 0.564 & 129.000 & 58.387 \\
DTD & \textbf{Base} & \textbf{71.738} & \textbf{1.915} & \textbf{218.400} & \textbf{69.984} \\
& 10 MN & 71.223 & 1.083 & 139.400 & 26.557 \\
& 10 M & 70.626 & 1.242 & 133.667 & 35.698 \\
 \hline
& 10 M & 95.584 & 0.581 & 113.800 & 14.822 \\
Brain Tumor & 25 M & 95.312 & 1.434 & 115.167 & 44.640 \\
& 10 MN & 95.295 & 1.370 & 91.250 & 48.417 \\
& \textbf{Base} & \textbf{95.082} & \textbf{1.629} & \textbf{117.500} & \textbf{40.806} \\
 \hline
 & 10 MN  & 89.733 & 0.281 & 125.000 & 74.579 \\
 Fruit 25 & 5 MN  & 89.520 & 0.000 & 183.000 & 26.870 \\
 & 25 M  & 89.200 & 0.339 & 147.000 & 7.071 \\
 & 10 M  & 89.147 & 0.521 & 126.500 & 45.822 \\
 & \textbf{Base} & \textbf{88.813} & \textbf{0.519} & \textbf{134.000} & \textbf{41.905} \\
\hline
\end{tabular}
\end{table}

On the DTD dataset, the 5 $M$ experimental case showed a marginal increase of around 0.3\% with a p-value of $0.38209$ when compared to the base case. The increase in convergence rate is significant at $p<0.05$ with a p-value of $0.03005$. The number of epochs to convergence decreased from an average of $218$ epochs to around $130$ epochs. This convergence gain is consistent across all $M$ and $MN$ experimental cases, representing a clear efficiency improvement with minimal loss in final testing accuracy. 

For the Brain Tumor dataset, plasticity restoration yielded improvements in both accuracy and convergence speed. The 10 $M$ case improved test accuracy by 0.5\% while reducing epochs to convergence by 5 epochs with p-values of $0.35942$ and $0.23885$ respectively. In the Fruit25 experiments, all $M$ and $MN$ outperformed the baseline, with the largest improvement in test accuracy being around 0.9\% accompanied by a 9-epoch improvement in convergence speed. This is a statistically significant accuracy increase at $p<0.05$ with a p-value of $0.00964$. The epoch improvement is not significant at $p<0.05$ with a p-value of $0.3409$.

In general, the most pronounced gains were observed on Fruit25, with improvements in both accuracy and convergence speed. The efficiency gains for DTD were substantial, nearly halving training time. These results suggest that our method scales well with deeper CNN architectures like ResNet-50, and has strong potential to provide significant efficiency benefits, particularly on atypical datasets.

\subsection{ViT-B16}\label{sec:vit}
ViT-B16, a transformer-based architecture, has a large number of parameters, and therefore even small improvements in convergence or test accuracy are impactful. The empirical results for ViT-B16  are presented in Table~\ref{tab:ViT-B16}.

\begin{table}[!htb]
\centering
\caption{A comparison between plasticity-restoring experimental cases and the base case on the \textit{ViT-B16} model. The specific experimental cases are listed using the notation defined in Section \ref{sec:approach}. The results are ordered by accuracy, and the base case is indicated in bold.}\label{tab:ViT-B16}
\begin{tabular}{|c|c|l@{\ $\pm$\,}l|l@{\ $\pm$\,}r|}
\hline
Dataset & Experimental Case &  Accuracy & Std & Epochs & Std\\
 \hline
& 5 M & 71.809 & 0.862 & 95.600 & 15.010 \\
DTD & \textbf{Base} & \textbf{71.649} & \textbf{0.887} & \textbf{97.800} & \textbf{18.185} \\
& 5 MN & 71.596 & 0.563 & 89.400 & 5.814 \\
& 25 MN & 71.241 & 0.437 & 90.000 & 6.745 \\
  \hline
& 25 M & 98.001 & 0.390 & 70.100 & 17.559 \\
Brain Tumor & 5 MN & 97.918 & 0.415 & 65.200 & 15.959 \\
& 10 M & 97.890 & 0.286 & 65.600 & 11.459 \\
& \textbf{Base} & \textbf{97.652} & \textbf{0.458} & \textbf{65.600} & \textbf{19.867} \\
  \hline
& 25 M & 90.700 & 0.528 & 78.200 & 21.936 \\
Fruit 25 & 5 M & 90.480 & 0.577 & 88.200 & 5.975 \\
& 25 MN & 90.440 & 0.273 & 75.200 & 20.067 \\
& \textbf{Base} & \textbf{89.904} & \textbf{0.677} & \textbf{74.400} & \textbf{12.341} \\
\hline
\end{tabular}
\end{table}

On the DTD dataset, a modest increase in test accuracy of approximately 0.2\% is observed, with no significant change in convergence. The second best experimental case, 5 $MN$, showed a more significant 8 epoch faster convergence at the cost of only a 0.1\% decrease in test accuracy. This gain in epoch rate is significant at $p<0.05$ with a p-value of $0.04363$. For the Brain Tumor dataset, accuracy improvements are more pronounced, with an increase of around 0.4\% with no significant change in convergence. Four experimental cases outperformed the baseline in accuracy, all while maintaining comparable convergence speed. All $M$ and $MN$ experimental cases outperformed the baseline on the Fruit25 dataset, with a maximum accuracy gain of about 0.8\%. Most experiments showed convergence rates similar to the baseline, except for the 5 $M$ experimental case, which required 14 more epochs. The p-values for the increase in accuracy are $0.04457$, $0.04093$, and $0.06426$ for DTD, Brain Tumor and Fruit respectively, which shows a significant improvement at $p<0.05$ for DTD and Brain Tumor. 

Collectively, these results are promising for ViT models. Our experiments demonstrate small but consistent improvements in test accuracy across all datasets, supporting our hypothesis that selective weight reinitialization can effectively re-enable learning plasticity in large transformer architectures.

\section{Critical Analysis}\label{sec:analysis}
Overall, selective reinitialization evidently restores a degree of neural plasticity and yields statistically significant empirical gains in both accuracy and training efficiency across diverse architectures and datasets. The remainder of this section examines the conditions under which the method is most effective by evaluating the effect of different architectures, datasets and reinitialization functions on the observed performance. 

\paragraph{Influence of Architecture:}
Restoring plasticity proved effective across all evaluated architectures. Convolutional models (SimpleNet and ResNet-50) exhibited the largest relative improvements, with SimpleNet achieving the highest single test accuracy gain, and ResNet-50 the greatest reduction in convergence time (Table \ref{tab:Best}). Our method demonstrates more consistent improvements in test accuracy on larger architectures, such as ResNet-50 and ViT-B16, and outperforms the baseline across all datasets in such cases. In contrast, our method is outperformed by the baseline on the Brain Tumor dataset using the SimpleNet architecture. This is an indication that smaller models may be more sensitive to dataset-specific characteristics. Specifically, the strong performance of the base case can be attributed to the Brain Tumor dataset having only four classes, which, given the smaller number of parameters, allows the smaller model to build a stronger understanding of each class. Our method generally improves convergence on the CNN architectures, whereas minimal convergence change is observed when running on the ViT architecture.

\begin{table}[!tb] 
\centering
\caption{Best runs across all datasets and architectures, presented as tuples. The first element in the tuple is the gain or loss in the best accuracy for a model, and the second element is the corresponding gain or loss in epochs for that best experimental case. Tuples with both a gain in plasticity and decrease in epoch (improved convergence rate) are indicated in bold.}\label{tab:Best}
\begin{tabular}{|l|l|l|l|}
\hline
Dataset & SimpleNet &  ResNet-50 & ViT-B16\\
\hline
DTD & $[2.4;+7]$ & $\textbf{[0.3;-90]}$ & $\textbf{[0.15;-2]}$   \\
Brain Tumor &  $[-0.7;-62]$ & $\textbf{[0.5;-5]}$ & $[0.4;+5]$  \\
Fruit25 & $\textbf{[0.5;-18]}$ & $\textbf{[0.9;-9]}$ & $[0.7;+4]$  \\
\hline
\end{tabular}
\end{table}
 
\paragraph{Influence of Dataset:}
The fine-tuning dataset strongly influenced the effectiveness of our method. The largest average accuracy gains occurred on Fruit25, likely due to both its relatively large size (compared to the other datasets) and its visual similarity to ImageNet. This suggests that our approach preserves useful pretrained representations while also re-enabling plasticity for downstream adaptation. On more atypical datasets, such as DTD and Brain Tumor, our method also enhanced adaptability, achieving the most substantial improvements on DTD despite its limited 96 training samples per class. By contrast, Brain Tumor offered more samples per class ($\sim1400$), and gains in accuracy and convergence were less pronounced. This is expected, and shows that additional data is more influential than the re-enabling of plasticity. Another strong correlation typically exists between training duration and final accuracy, and our results exceed what would be expected from runtime effects alone. As can be seen in Table \ref{tab:Best}, several experimental cases improved both convergence rate and test accuracy simultaneously, supporting the view that selective reinitialization directly improves model adaptability across datasets.

\begin{figure}
    \centering
       \begin{subfigure}[b]{0.71\textwidth} 
        \centering
        \includegraphics[width=\linewidth]{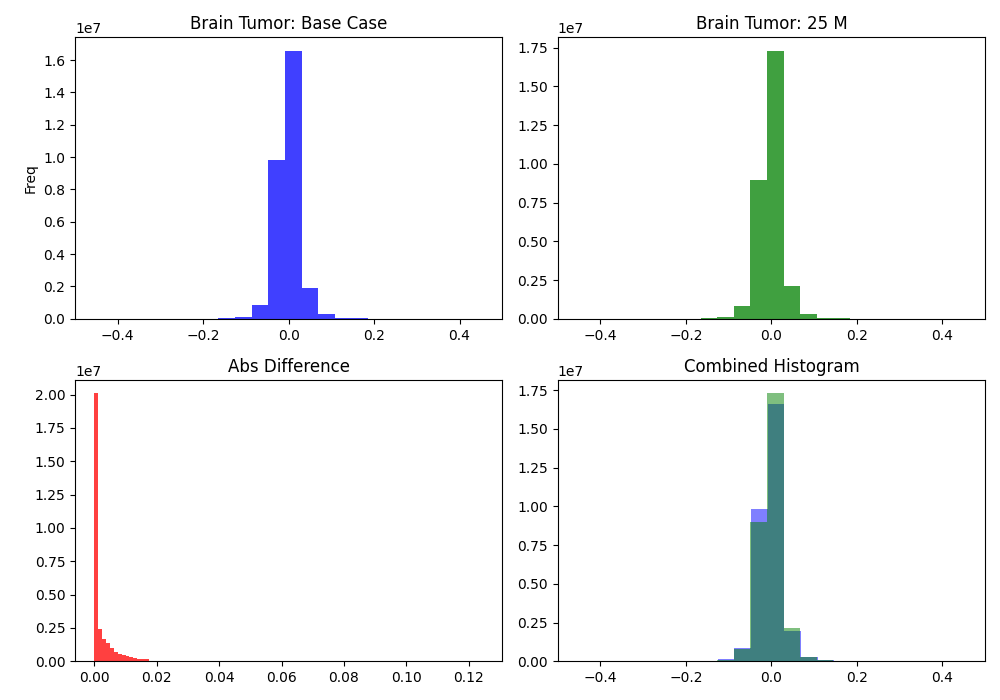} 
        \caption{25 $M$}
        \label{fig:subfigA}
    \end{subfigure}
    \begin{subfigure}[b]{0.71\textwidth}
        \centering
        \includegraphics[width=\linewidth]{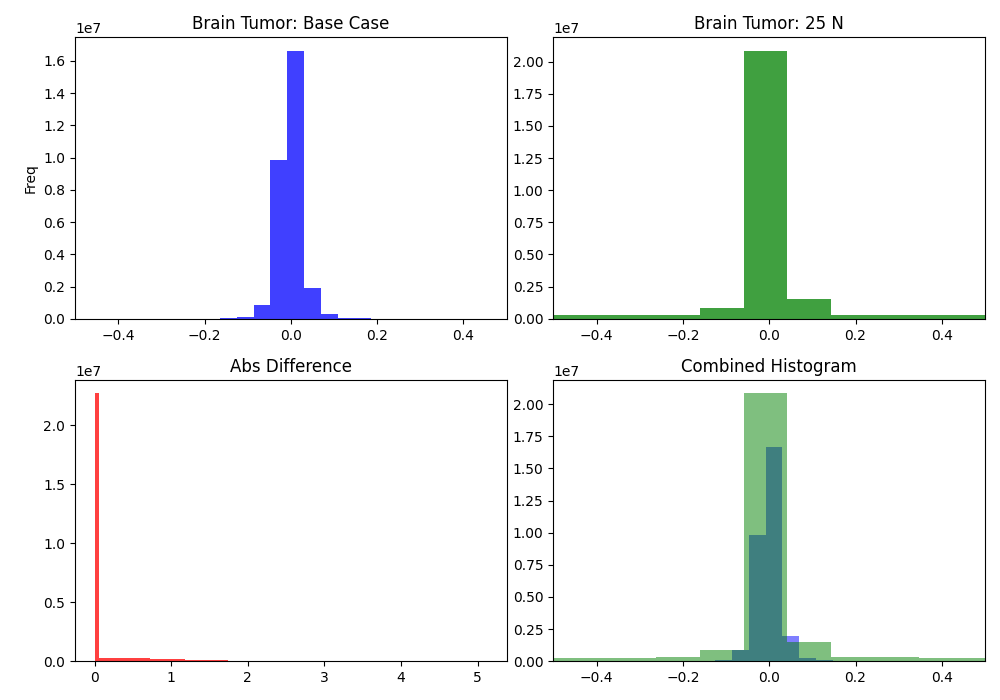}
        \caption{25 $N$}
        \label{fig:subfigB}
    \end{subfigure}

    \caption{Comparison between 25 $M$ (a) and 25 $N$ (b) experimental cases. The blue histogram is the total weight distribution after training on the base case. The green histogram represents the total weight distribution after fine-tuning in the experimental cases. The red histogram shows the absolute difference between the blue and green histograms and the final histogram overlays the blue and green histograms for easy comparison between base and experimental cases.}
    \label{fig:weight_dist}
\end{figure}

\paragraph{Effect of Reinitialization Function:}
Across all experiments, the most effective reinitialization strategies were the $M$ and $MN$ functions. In contrast, less suitable distributions (e.g., $N$ and $NS$) consistently degraded performance. This performance degradation can be attributed to a large out-of-distribution change to the initial weights. The effect of this as on the final weight distribution of the model can be observed under \textit{Brain Tumor: 25 N} in Figure~\ref{fig:weight_dist}. Figure~\ref{fig:weight_dist} illustrates the difference between the weight distributions before and after reinitialization for both the 25 $M$ and 25 $N$ reinitialization function on ResNet-50. As seen in Figure~\ref{fig:weight_dist}, the most impactful modifications to the weight distribution were modest in scale (25 $M$) in comparison to the big changes made by 25 $N$. This is consistent with prior findings that large initial weights can slow down training~\citeyearpar{dohare2024loss} and degrade final test accuracy. For less effective reinitialization functions, performance often improved as the proportion of reinitialized weights decreased, suggesting that excessive disruption can harm transfer learning. Conversely, for the more effective reinitialization functions, larger proportions of reinitialized weights tended to yield greater accuracy gains and/or faster convergence. Optimal proportions varied per dataset: DTD typically benefited most from 5\% reinitialization, while Fruit25 and Brain Tumor achieved peak performance at 10\%–25\%. 

\paragraph{Summary:}
Together, these results highlight that the effectiveness of selective weight reinitialization depends on both architectural characteristics and dataset properties. Smaller CNNs such as SimpleNet can exhibit large but less consistent gains, while deeper architectures, such as ResNet-50 and ViT-B16, tend to deliver more stable improvements across datasets. Although dataset similarity to the pretraining domain affects fine-tuning performance, it does not reduce the effectiveness of the proposed algorithm. The choice of reinitialization function is critical, with $M$ and $MN$ consistently outperforming alternatives and demonstrating that small adjustments before fine-tuning can have a meaningful effect. Importantly, the results reported in this paper are aligned with related work in transfer learning and parameter-efficient fine-tuning.

\subsection{Comparison to Other Methods}
We compare the performance improvements achieved in this work, applied before finetuning, with those reported in related studies (Section~\ref{sec:related}), applied during finetuning. Specifically, we consider the approaches of Li et al.~\citeyearpar{xuhong2018explicit} and Chen et al.~\citeyearpar{chen2024conv}.  

Li et al.~\citeyearpar{xuhong2018explicit} demonstrate accuracy gains of 1–4\% through the use of an explicit inductive bias, though without any corresponding efficiency improvement. Our method yields accuracy gains of 0.5–2\%, with modest efficiency improvements in certain cases. Chen et al.~\citeyearpar{chen2024conv} also report only marginal improvements in test accuracy, but with substantial efficiency benefits: for example, ConvAdapter requires only 3.5\% of the learnable parameters needed for full fine-tuning of ResNet-50. The efficiency improvements observed in our approach are therefore not directly comparable to those of ConvAdapter.  

We note that our approach can be combined with the approaches of Li et al.~\citeyearpar{xuhong2018explicit} and Chen et al.~\citeyearpar{chen2024conv}, with the potential of stacking the benefits.

\section{Conclusion}\label{sec:conclusion}
This paper proposed a selective weight reinitialization method to mitigate loss of plasticity during transfer learning by re-enabling learning plasticity before fine-tuning. By selectively reinitializing a subset of inactive neural network weights, we observed improved fine-tuning accuracy and faster convergence across diverse datasets. Importantly, no experimental case showed that the baseline significantly outperformed selective reinitialization, indicating the robustness of the method and the potential for broad applicability in fine-tuning tasks. For most cases, we recommend reinitializing 10\% of the weights using the mean reinitialization strategy. In situations where the fine-tuning dataset is extremely atypical, we recommend a reinitialization proportion of 5\% for the best results. 

Future research should investigate more advanced reinitialization techniques and adaptive approaches to further address plasticity loss. Extending this methodology to large language models and bigger datasets also presents a promising direction for enhancing fine-tuning efficiency. Furthermore, the effect of hyper parameter choices (learning rates, batch sizes, etc.) on the sensitivity of the results remains to be explored.

\bibliographystyle{unsrtnat}
\bibliography{references}  






\end{document}